\documentclass{egpubl}
\usepackage{egsr2021_DLOnlyTrack}
 
\WsPaper           

\usepackage[T1]{fontenc}
\usepackage{times}
\usepackage{ifpdf}
\usepackage{dfadobe}
\usepackage{amsmath}
\usepackage{amssymb}
\usepackage{pifont}
\biberVersion
\BibtexOrBiblatex
\usepackage[backend=biber,bibstyle=EG,citestyle=alphabetic,backref=true]{biblatex} 
\electronicVersion
\PrintedOrElectronic
\ifpdf \usepackage[pdftex]{graphicx} \pdfcompresslevel=9
\else \usepackage[dvips]{graphicx} \fi

\usepackage{egweblnk} 

\begin{document}

\definecolor{gray}{rgb}{0.5,0.5,0.5}
\definecolor{purple}{rgb}{0.7,0.3,0.7}
\definecolor{blue}{rgb}{0,0,1}
\definecolor{darkblue}{rgb}{0,0,0.6}
\definecolor{orange}{rgb}{1,.5,0} 
\definecolor{red}{rgb}{1,0,0} 

\newcommand{\todo}[1]{{\textcolor{red}{TODO: #1}}}
\newcommand{\changed}[1]{#1}
\newcommand{\de}{\mathrm{d}}
\newcommand{\fig}[1]{Figure~\ref{#1}}

\newcommand{\Mat}[1]{{\ensuremath{\mathbf{\uppercase{#1}}}}} 
\newcommand{\Vect}[1]{{\ensuremath{\mathbf{\lowercase{#1}}}}} 
\newcommand{\Vari}[1]{{\ensuremath{\mathrm{\lowercase{#1}}}}} 
\newcommand{\Id}{\mathbb{I}} 
\newcommand{\Diag}[1]{\operatorname{diag}\left({ #1 }\right)} 
\newcommand{\Opt}[1]{{#1}_{\text{opt}}} 
\newcommand{\minimize}[1] {\underset{{#1}}{\operatorname{min}} \: \: } 
\newcommand{\maximize}[1] {\underset{{#1}}{\operatorname{max}} \: \: } 
\newcommand{\argmin}[1] {\underset{{#1}}{\operatorname{argmin}} \: \: }
\newcommand{\argmax}[1] {\underset{{#1}}{\operatorname{argmax}} \: \: } 

\newcommand{\pos}{\Vect{x}}
\newcommand{\dir}{\omega}
\newcommand{\tgtdir}{\omega_t}
\newcommand{\envmap}{L}
\newcommand{\envmapmat}{\Mat{L}}
\newcommand{\lt}{T}
\newcommand{\ltmat}{\Mat{T}}
\newcommand{\radiance}{L}
\newcommand{\inradiance}{L_{i}}
\newcommand{\outradiance}{L_{r}}
\newcommand{\vis}{\Mat{V}}
\newcommand{\brdf}{\Mat{\rho}}
\newcommand{\norm}{\Vect{n}}

\newcommand{\density}{\Vect{\sigma}}
\newcommand{\image}{\Mat{I}}
\newcommand{\mask}{\Mat{M}}

\newcommand{\cnn}{\textbf{CNN}}
\newcommand{\mlp}{\mathcal{F}}
\newcommand{\imagefeature}[1]{\Mat{F}_{#1}}
\newcommand{\geometryfeature}[1]{\Mat{G}_{#1}}

\newcommand{\weight}[2]{\Mat{W}_{#1}^{\textbf{(#2)}}}
\newcommand{\envmapweight}[1]{\Mat{W}_{#1}^{\envmapmat}}
\newcommand{\geometryweight}[1]{\Mat{W}_{#1}^{G}}
\newcommand{\blendingweight}[1]{\Mat{W}_{#1}^{B}}
\newcommand{\rot}{\Mat{R}}
\newcommand{\rotate}{\mathcal{R}}

\newcommand{\camnum}{N}

\newcommand{\response}{\Mat{R}}
\newcommand{\dirimage}{\Mat{\Theta}}
\newcommand{\confimage}{\Mat{W}}
\newcommand{\vectorize}[1]{\textrm{Vec}({#1})}
\newcommand{\lightcode}{\Mat{L}}

\newcommand{\pixel}{p}
\newcommand{\segment}{S}
\newcommand{\segmentpos}{S_{pos}}
\newcommand{\segmentdir}{S_{dir}}


\newcommand{\IGNORE}[1]{}

\definecolor{MyDarkBlue}{rgb}{0,0.08,1}
\definecolor{MyDarkGreen}{rgb}{0.02,0.6,0.02}
\definecolor{MyDarkRed}{rgb}{0.8,0.02,0.02}
\definecolor{MyDarkOrange}{rgb}{0.70,0.35,0.02}
\definecolor{MyPurple}{rgb}{0.43,0,1.}
\definecolor{MyRed}{rgb}{1.0,0.0,0.0}
\definecolor{MyGold}{rgb}{0.75,0.6,0.12}
\definecolor{MyDarkgray}{rgb}{0.66, 0.66, 0.66}

\newcommand{\tc}[1]{{\color{MyRed}\textbf{\emph{Tiancheng: #1}}}}
\newcommand{\KL}[1]{{\color{MyDarkOrange}\textbf{\emph{Kai-En: #1}}}}
\newcommand{\ravi}[1]{\textcolor{MyDarkGreen}{[Ravi: #1]}}


    \title{NeLF: Neural Light-transport Field for\\ Portrait View Synthesis and Relighting}
    
    \author[T. Sun, K. Lin, S. Bi, Z. Xu, R. Ramamoorthi]
    {\parbox{\textwidth}{\centering
    Tiancheng Sun$^{1*}$,
    Kai-En Lin$^{1*}$,
    Sai Bi$^{2}$,
    Zexiang Xu$^{2}$,
    Ravi Ramamoorthi$^{1}$
    }
            \\
    {\parbox{\textwidth}{\centering
    $^1$University of California, San Diego, 
    $^2$Adobe Research\\
    $^*$Equal contribution
           }
    }
    }
    
    \teaser{
        \includegraphics[width=\linewidth]{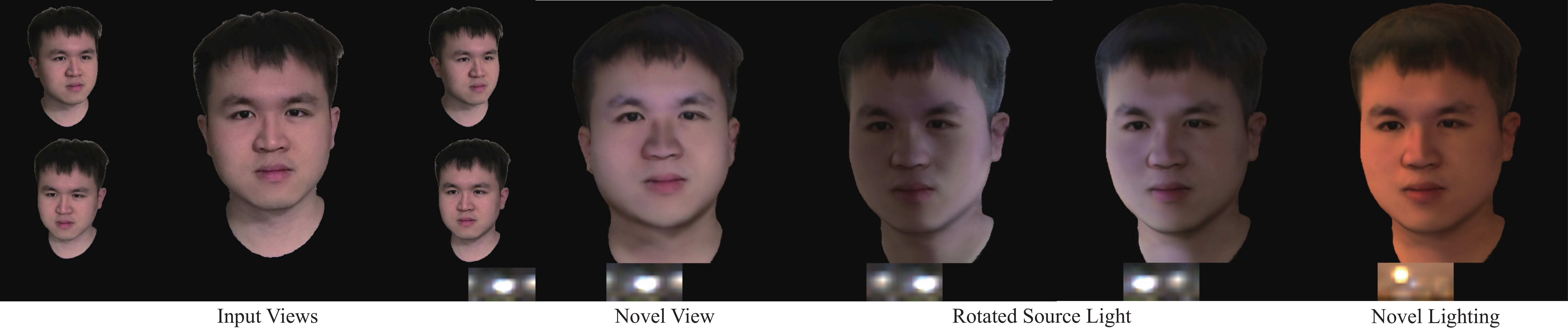}
        \centering
        \caption{We introduce neural light-transport field (NeLF), which learns to infer the light-transport and volume density from a sparse set of input views. NeLF enables joint relighting and view synthesis of real portraits from only five input images.}
        \label{fig:teaser}
    }
    
    \maketitle

    \begin{abstract}

Human portraits exhibit various appearances when observed from different views under different lighting conditions. We can easily imagine how the face will look like in another setup, but computer algorithms still fail on this problem given limited observations.
To this end, we present a system for portrait view synthesis and relighting: given multiple portraits, we use a neural network to predict the light-transport field in 3D space, and from the predicted Neural Light-transport Field (NeLF) produce a portrait from a new camera view under a new environmental lighting.
Our system is trained on a large number of synthetic models, and can generalize to different synthetic and real portraits under various lighting conditions.
Our method achieves simultaneous view synthesis and relighting given multi-view portraits as the input, and achieves state-of-the-art results.

\end{abstract}
    \section{Introduction}
Digitizing human portraits from natural portrait images and re-synthesizing novel images under new lighting and viewpoints is a long-standing graphics and vision problem with many applications like computational photography and video conferencing.
Recently, NeRF \cite{mildenhall2020nerf} has initiated significant progress in neural rendering for photo-realistic image synthesis. Various radiance field extensions~\cite{nerv2021,martinbrualla2020nerfw,bi2020neural,Liu20neurips_sparse_nerf} have been presented and many of the concurrent works~\cite{Gafni20arxiv_DNRF,lombardi2021mixture,Park20arxiv_nerfies} aim to address problems in human portrait capture.

However, previous NeRF-based portrait capture methods usually require capturing a large number of images and an expensive per-scene optimization process. Besides, most works consider only view synthesis tasks and do not support relighting. 
In general, a relightable portrait representation cannot simply be a radiance field that bakes in the original lighting condition; it instead requires explicitly disentangling and modeling lighting and portrait appearance in the image captures, which is a highly challenging problem.

In this paper, we propose a novel neural rendering approach that can jointly estimate the geometry and appearance of a human portrait and the lighting environment the human stands in, from only a sparse set of input images.
We present a novel deep neural network that can generalize across scenes to regress a \emph{neural light transport field}, i.e. NeLF, from portrait images. At an arbitrary 3D point, this neural light transport field outputs the volume density and light transport coefficients (unlike the view-dependent radiance in NeRF), which linearly explains the portrait appearance under distant illumination conditions represented by environment maps, similar to the classic light transport functions in image-based relighting \cite{debevec2000lightstage} and precomputed radiance transfer \cite{sloan2002precomputed,ng2003allfreq}. With this novel representation, our approach enables high-quality portrait relighting and view synthesis done simultaneously, as shown in Fig.~\ref{fig:teaser}.

Our approach incorporates classic light transport functions into neural volumetric rendering. 
In particular, we first use a UNet-like CNN as a feature extractor to convert each input portrait image to a neural feature map that encodes per-view pixel-wise portrait geometry and appearance.
For an arbitrary 3D point, we fetch multi-view neural features from its projections in the feature maps and use MLPs to regress volume density and radiance from the features to enable differentiable ray marching.
As opposed to earlier works \cite{yu2020pixelnerf,wang2021ibrnet} that directly output radiance per shading point on marching rays, we propose to first regress a \emph{light transport vector} -- that can linearly compute the radiance under any novel lighting conditions --  to enable relighting for neural volumetric rendering.
In addition, we apply a sub-network to estimate
the lighting as environment maps from the bottleneck features of the CNN feature extractor, contributing to the disentanglement of portrait appearance and light effects in the original images.
Our approach jointly estimates portrait geometry (as volume density), appearance (as light transport vectors) and lighting conditions (as environment maps) from portrait images, and can regress final ray colors under arbitrary novel viewpoints and lighting via differentiable ray marching.

Acquiring a real portrait dataset under different lighting conditions is a notoriously challenging task and traditionally requires a sophisticated light stage~\cite{debevec2000lightstage} that is not easily accessible for most researchers.
In order to train our network with practical data, we instead utilize rendered images of human head models with different views and environment maps.
We render our training and validation sets using the reconstructed 3D models of real human heads from a public dataset, FaceScape~\cite{yang2020facescape}, leading to realistic renderings close to real portraits.
In addition, we propose an effective domain adaptation module to enhance the generalizability of our network to real captured images.
In particular, we apply additional CNN layers, appending to our 
CNN feature extractor, to regress the original input image sent to the feature extractor.
We train this regression module using a large number of real portrait images in CelebAMask-HQ~\cite{CelebAMask-HQ}, which effectively regularizes our feature extractor along with our full network to understand diverse real portrait appearance.
This regularization module effectively improves our rendering quality on real portrait images (compared with the rightmost image in Fig.~\ref{fig:ablation}).

We train our entire network from end to end with a final loss that combines the lighting estimation loss, domain regularization loss, and governing rendering losses with multiple combinations of various lighting and viewing conditions.
This leads to our final generalizable network that can synthesize realistic portrait images under novel viewpoints and lighting conditions from only five input images.
We demonstrate that, when evaluating on the rendered validation set, our approach can produce smooth and realistic relighting and view synthesis results that are very close to the ground truth; ours can qualitatively and quantitatively outperform baseline solutions that run state-of-the-art view synthesis \cite{wang2021ibrnet} and relighting \cite{sun2019single} techniques in a sequence (see Fig.~\ref{fig:comp}).
We also demonstrate that our approach can achieve photo-realistic renderings from captured real portrait images, significantly better than the comparison methods (see Fig.~\ref{fig:real}).

We summarize our contributions as follows:
\begin{itemize}
    \item a novel neural representation that models scene appearance as light transport functions and enables relighting for neural volumetric rendering (Sec.~\ref{sec:lt}, Sec.~\ref{sec:nelf});
    \item a domain adaptation module to enhance the generalizability of the network trained on rendered images (Sec.~\ref{sec:aug});
    \item realistic practical rendering results of joint relighting and view synthesis of real portraits from only five captured images (Sec.~\ref{sec:real_portrait}, Fig.~\ref{fig:teaser}, Fig.~\ref{fig:real}).
\end{itemize}

    \section{Related Work}

\subsection{Portrait Appearance}
There has been extensive research work in capturing and modeling human portraits \cite{blanz1999morphable,debevec2000lightstage,kemelmacher20103d,tran2018nonlinear,sengupta2018sfsnet,zollhofer2018state}.
Various morphable face models have been presented \cite{tran2018nonlinear,garrido2014automatic,weise2011realtime}, mainly aiming to achieve facial animation and reenactment.
Our focus is to capture human portraits and synthesize photo-realistic images.
While generative models \cite{goodfellow2014generative,karras2017progressive,karras2019style} can produce realistic images, they cannot be easily applied for capture and reconstruction, which aims to generate images of specific captured real human heads with given viewpoints and lighting conditions.
While previous work can do so by reconstructing meshes with simple reflectance models from portrait images \cite{barron2014shape,sengupta2018sfsnet}, their rendering quality is often limited. We instead leverage the recent neural rendering techniques~\cite{tewari2020state}, leading to photo-realistic portrait renderings for joint view synthesis and relighting.

\subsection{Relighting}
Many image-based relighting papers have been presented \cite{debevec2000lightstage,matusik2004progressively,peers2009compressive,ren2015image}; they leverage the linearity of light transport and introduce various techniques to acquire the light transport function, whose discrete form is a light transport matrix that consists of per-pixel light transport vectors at a fixed viewpoint.
Inspired by these prior works, we also model linear light transport functions; instead of predicting per-pixel coefficients, we predict light transport at a point in the 3D space, enabling both view synthesis and relighting.

\begin{figure*}
    \centering
    \includegraphics[width=1.0\textwidth]{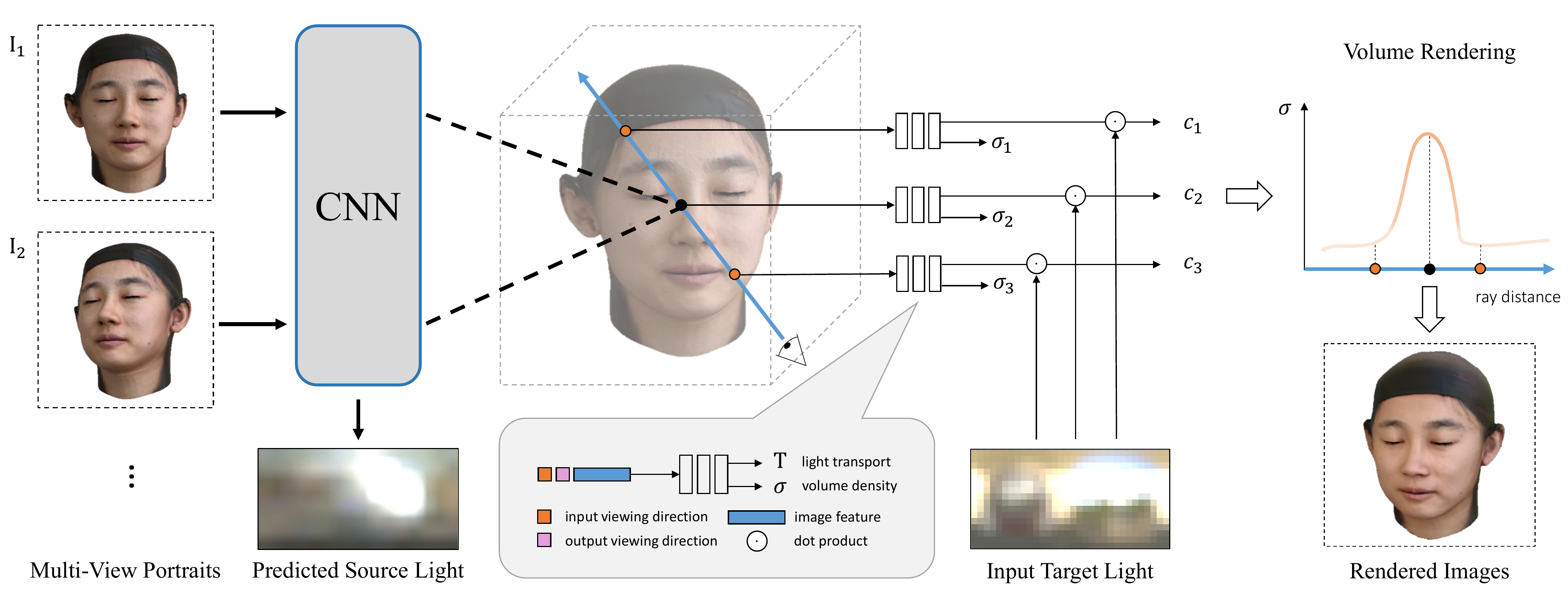}
    \caption{\label{fig:overview}Overview of our method. The proposed algorithm takes multi-view portraits as input and predicts the source environment map, light-transport and volume density at a query point. We then use the predicted light-transport and volume density to perform the joint task of view synthesis and relighting.}
\end{figure*}

Recently, deep learning techniques have been introduced to address the relighting problem \cite{xu2018deep}.
Many deep learning works are specifically designed to relight human portraits~\cite{sun2019single,zhou2019deep,meka2019deep,nestmeyer2020learning,sun2020light}; however most works focus on relighting at a fixed viewpoint and cannot change the viewpoint.
While other works can jointly do relighting and view synthesis \cite{bi2020deep3d,meka2020deep,gao2020deferred,zhang2021neural}, they require complex capture setups to acquire a large number images under controlled lighting.
In contrast, our approach enables realistic relighting and view synthesis for human portraits using only a sparse set of input images under natural illumination.


\subsection{View Synthesis}
View synthesis has been studied by the computer graphics and vision community for decades.
Early work~\cite{chen1993view,mcmillan1995plenoptic} utilizes view interpolation to render novel views from neighboring source views.
Other classical methods, including light fields~\cite{levoy1996light, gortler1996lumigraph} and image-based rendering~\cite{debevec1996modeling,buehler2001unstructured,sinha2009piecewise}, have also been proposed to address view synthesis.
Recently, deep learning methods~\cite{zhou2018stereo,mildenhall2019local,xu2019deep,flynn2016deepstereo} have become dominant in this field of work.
Learning-based methods have proven to be more expressive and they are able to represent various complex scenes with challenging visual effects.
One recent work \cite{xu2020deep} combines morphable face models and generative techniques to reconstruct 3D portraits for rendering with changing viewpoints.
However, it is non-trivial to extend these view synthesis techniques to support relighting at the same time.
We propose a novel neural rendering approach that can simultaneously do relighting and view synthesis.




\subsection{Neural Rendering}
In addition to the aforementioned view synthesis methods, an exciting advancement is neural radiance field (NeRF)~\cite{mildenhall2020nerf}, which encodes a 3D scene in a compact 5D continuous radiance field function represented by a multi-layer perceptron (MLP) and renders the radiance field using differentiable volume rendering.
Concurrent works have extended NeRF to render human portraits~\cite{Park20arxiv_nerfies,wang2020learning,lombardi2021mixture}; however most of them rely on overfitting the network to a single target as is done in the original NeRF, which does not generalize to other unseen portraits.
Other recent works \cite{yu2020pixelnerf,wang2021ibrnet} leverage CNNs to perform per-view radiance field reasoning, leading to a generalizable neural model for view synthesis. 
Our approach is inspired by these CNN-based radiance field estimation techniques;
we introduce novel light transport estimation modules in the per-view scene reasoning, enabling relighting in the neural volumetric rendering process.

Previous neural rendering methods have also achieved relighting. Some methods leverage controlled lighting to achieve reflectance estimation with per-scene optimization \cite{bi2020deep,bi2020neural}. Our approach instead supports relighting under natural illumination, jointly estimating lighting and lighting transport functions.
Other concurrent works \cite{nerv2021,martinbrualla2020nerfw} can also change the lighting conditions but still rely on per-scene optimization with a large number of images. 
Our approach achieves a generalizable neural network that learns specific human portrait shape and appearance priors from large training datesets.
Additionally, our approach enables efficient portrait relighting and view synthesis from only a sparse set of input images.

    \section{Method}

In this paper, we present Neural Light-transport Field (NeLF) to solve the problem of simultaneous portrait view synthesis and relighting.
Our system takes as input a small set of 5 images of an unseen human face, which are captured roughly from the frontal view of the portrait. We assume the captured portraits are lit by the same distant light, which can be modeled by an environment map.

Given the captured images as well as the corresponding camera parameters, we produce a volumetric field of \textit{light-transport} (Sec.~\ref{sec:lt}). Each point in the 3D scene has a volume density $\density$, and a light-transport $\lt$.
The light-transport is a vector whose dot-product with the global environment map produces the outgoing radiance at the 3D point.
The predicted neural light-transport field (NeLF) enables us to perform view synthesis and relighting on the captured portraits. Given a new camera view and a new lighting environment, we can use the predicted light-transport to compute the outgoing radiance of each 3D point, and use a volume rendering algorithm~\cite{mildenhall2020nerf} to render novel views of the captured human face under the new lighting.

\begin{figure*}[t]
    \centering
    \includegraphics[width=1.0\textwidth]{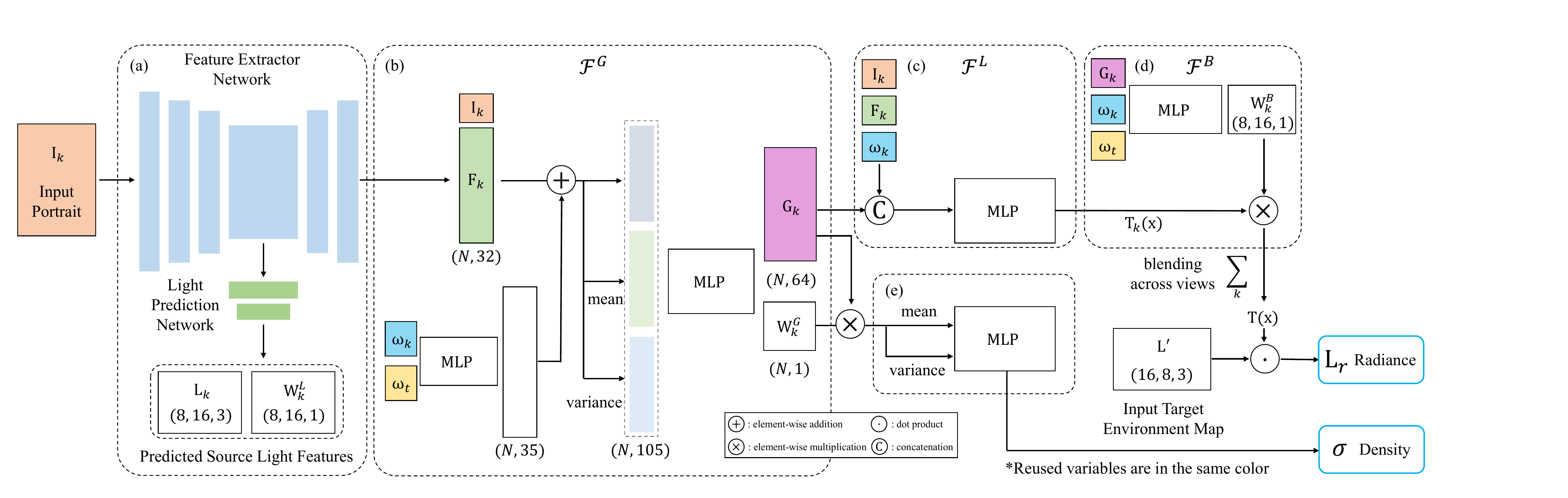}
    \caption{\label{fig:network}Detailed network structures of our proposed algorithm. (a) For an input image $\image_k$, we first extract its lighting feature $\envmapmat_k$, weights $\envmapweight{k}$ and image feature $\imagefeature{k}$ (Sec.~\ref{sec:cnn}). (b) Then, as described in Sec.~\ref{sec:mlp}, we aggregate features from different views to extract the geometry feature, $\geometryfeature{k}$, and weights $\geometryweight{k}$. From here, our network diverges into two paths: the first part (c) predicts the per-view light transports $\ltmat_k$ using a per-view MLP $\mathcal{F}^L$ and then (d) aggregates them with a blending MLP $\mathcal{F}^B$ to generate the full light transport $\ltmat$; the second part (e) predicts the volume density $\density$ using an MLP from the multi-view statistics of $\geometryfeature{k}$ and  $\geometryweight{k}$.}
\end{figure*}

We show an overview of our method in Fig.~\ref{fig:overview} and a detailed illustration in Fig.~\ref{fig:network}.
In Sec.~\ref{sec:lt}, we briefly review the mathematical definition of the light-transport.
Section~\ref{sec:nelf} describes how we predict the volume density and the light-transport at each 3D point, and Sec.~\ref{sec:render} explains our detailed volume rendering algorithm.
We show our novel domain adaptation module in Sec.~\ref{sec:aug}.
Finally, the implementation details are included in Sec.~\ref{sec:detail}.

\subsection{Light Transport}~\label{sec:lt}
Light transport of a 3D point describes the relationship between its radiance and the global lighting environment. The outgoing radiance $\outradiance$ of each 3D point $\pos$ can be computed according to the rendering equation~\cite{kajiya1986rendering}:
\begin{equation}\label{equ:render}
    \outradiance(\pos, \dir_o) = \int_{\dir_i} \brdf(\pos, \dir_i, \dir_o) (\norm \cdot \dir_i) \inradiance(\pos, \dir_i)\de\dir_i,
\end{equation}
where $\brdf$ is the BSDF function, $\norm$ is the normal direction at the 3D point, and $\dir_i$, $\dir_o$ are the incoming and outgoing ray directions, respectively. Traditional path tracing computes the radiance $\inradiance$ recursively until the ray hits the global environment $\envmap$. However, after unravelling the recursion, the radiance has a linear relationship with the environment map:
\begin{equation}\label{equ:linear}
    \inradiance(\pos, \dir_i) \propto \envmap(\dir_i').
\end{equation}
Here, the incoming radiance $\inradiance$ of the point $\pos$ at each incoming direction $\dir_i$ is proportional to the corresponding global environment light $\envmap(\dir_i')$ at each direction $\dir_i'$. Notice that this linear relation has modeled all the paths that come from the environment to this point, modeling effects caused by shadowing, inter-reflection, material properties of the ray bounces in between, etc. If we bring Eqn.~\ref{equ:linear} back to Eqn.~\ref{equ:render}, we can simplify Eqn.~\ref{equ:render} as a linear integration:
\begin{equation}\label{equ:lt}
    \outradiance(\pos, \dir_o) = \int_{\dir_i} \lt(\pos, \dir_o, \dir_i) \envmap(\dir_i)\de\dir_i.
\end{equation}
Here, the \textit{light transport} $\lt$ has already encoded all the complexity from the environmental light $\envmap$ to the outgoing radiance $\outradiance$ at $\pos$.


As shown in Eqn.~\ref{equ:lt}, given the light transport $\lt$, we can compute the outgoing radiance at each 3D point $\pos$ under a novel environment map $\envmap$ using an integration, and then efficiently render the scene under arbitrary lighting environments. This type of method is called Precomputed Radiance Transfer (PRT)~\cite{sloan2002precomputed}, which has been well studied in traditional computer graphics~\cite{tsai2006all,ng2003allfreq}.
In our method, we express the global lighting simply as a $8 \times 16$ environment map $\envmapmat$. We predict a light-transport vector $\ltmat$ at each 3D point $\pos$ given an input viewing direction. $\ltmat$ also shares the same size as the environment map $\envmapmat$. As a result, we can simplify Eqn.~\ref{equ:lt} to a dot product on each color channel:
\begin{equation}\label{equ:lt_calc}
    \Mat{L}_r = \ltmat \cdot \envmapmat.
\end{equation}
This is similar to the light transport matrix in image-based relighting \cite{debevec2000lightstage}, which models per-pixel light transport. \changed{In our work, we compute the radiance at each 3D point following Eqn.~\ref{equ:lt_calc}, and later render the radiances into pixel values following the neural volumetric rendering equation in Eqn.~\ref{equ:vol_render}}.

\subsection{Neural Light-transport Field}~\label{sec:nelf}
Given $\camnum$ different views of a human face, we use a neural network to predict a volumetric light-transport field.
The key concept of our algorithm is to predict the light-transport vector $\ltmat$ of each 3D point by aggregating the information from the input portraits.
To this end, we utilize the ideas of the recent image-based rendering techniques~\cite{wang2021ibrnet} to predict the light-transport. Our system consists of two parts (see Fig.~\ref{fig:network}): a convolutional neural network (CNN) and several multi-layer perceptrons (MLP).
The CNN operates on the captured portraits to extract the image features and predict the source lighting condition.
On the other hand, the MLPs predict the light transport $\ltmat$ as well as the volume density $\density$ for each 3D point in the scene, using multiple features.
Then for any given target environment map $\envmapmat'$, we can simple calculate the color with Equ.~\ref{equ:lt_calc} and use it for volumetric rendering (Sec.~\ref{sec:render}).

\subsubsection{Image Feature Extraction and Source Light Prediction}~\label{sec:cnn}
We apply an U-Net style convolutional neural network~\cite{ronneberger2015u} to $\camnum$ captured images.
For an input portrait $\image_k$ at viewpoint $k$, the CNN extracts the image feature $\imagefeature{k}$, as well as the source lighting feature $\envmapmat_k$ and confidence weights $\envmapweight{k}$ (see Fig.~\ref{fig:network}(a)).
The size of image feature $\imagefeature{k}$ is half of the original input image $\image_k$, and it is later used for light transport prediction.

Each lighting feature consists of two parts: a predicted source environment map $\envmapmat_k$, and a per-direction confidence map $\envmapweight{k}$~\cite{sun2019single}.
We predict the confidence map $\envmapweight{k}$ for all cameras, as each camera covers only parts of the portrait, providing partial lighting information.
For example, it is less accurate to predict the light coming from the left by looking at the right side of the face.
To resolve this issue, we can merge the lighting predictions $\envmapmat_1, \envmapmat_2, ..., \envmapmat_N$ with their corresponding confidence maps $\envmapweight{1}, \envmapweight{2}, ..., \envmapweight{N}$ as weights. 

Nonetheless, another issue is that the network is unaware of the camera pose when predicting the lighting environment.
As a result, the network is only able to predict the lighting relative to each camera, \changed{and each predicted environment-map is defined in its corresponding camera coordinate system.}
In order to \changed{align multiple light predictions from different coordinate systems} and merge them into a global environment map, we define a rotation operator $\rotate$ that rotates them to a canonical world coordinate, provided camera extrinsics.
Finally, we compute the global lighting environment with a weighted average:
\begin{equation}
    \envmapmat = \frac{\sum_k\rotate(\envmapmat_k\odot\envmapweight{k})}{\sum_k\rotate(\envmapweight{k})},
\end{equation}
where $\odot$ means element-wise multiplication.

\subsubsection{Volume Density and Light-transport Prediction}~\label{sec:mlp}
We use multiple MLPs to predict the volume density $\density(\pos)$ and the light transport $\ltmat(\pos)$ at each point $\pos$ observed from the target camera $t$. 
We first project the point $\pos$ to all source cameras to acquire the corresponding image features $\imagefeature{k}(\pi_{k}(\pos))$, where $\pi_{k}$ denotes the projection to camera $k$.
We also compute source viewing direction $\dir_k$ and target viewing direction $\dir_t$.
The image features, together with $\dir_k$ and $\dir_t$, are fed into MLPs, $\mlp^{G}$, to extract the multiview-aware geometry feature $\geometryfeature{k}$ and a corresponding weight $\geometryweight{k}$ for each source view (see Fig.~\ref{fig:network}):
\begin{equation}
\geometryfeature{k}, \geometryweight{k} = \mlp^{G}\left(
    \dir_t, \{\dir_k\}_{k=1}^{\camnum}, \{\imagefeature{k}(\pi_{k}(\pos))\}_{k=1}^{\camnum}
\right).
\end{equation}
Shown in Fig.~\ref{fig:network}(b), the per-view geometry feature is extracted using a PointNet style MLP structure~\cite{qi2017pointnet}, using the per-element mean and variance of the feature as additional inputs (see Fig.~\ref{fig:network}).
This shares the idea of traditional image-based rendering~\cite{mcmillan1995plenoptic}: the projected image features from multiple cameras should be consistent around the actual object surface.
In our setup, $\mlp^{G}$ compares the input feature vector with its per-element mean and variance, and learns to assign more weight $\geometryweight{k}$ to the best matched view.
This geometry feature is used to predict the volume density $\density(\pos)$ of the query point using another MLP as shown in Fig.~\ref{fig:network} (e).

It is highly challenging to predict the light transport vector for the novel viewing direction. Therefore, we instead predict per-view light transports using an MLP $\mlp^{L}$ (Fig.~\ref{fig:network}(c)) and then blend them for the novel view (Fig.~\ref{fig:network}(d)). 
While we can apply $\mlp^{L}$ to directly regress the light transport vector, this can easily lead to the network memorizing the portrait appearance in the synthetic training set.
We propose to let $\mlp^{L}$ regress scales relative to the pixel colors; the predicted scales have the same dimensions as the light transport, and compute the light transport vector by multiplying by the pixel colors. This design effectively retains the high-frequency information in the original input images and enables better generalizability of our network to unseen real portraits (see Fig.~\ref{fig:ablation}).
In particular, the light-transport vector of the source view $k$ at point $\pos$ is computed by:
\begin{equation}~\label{equ:lt_pred}
    \ltmat_k(\pos) = \image_k(\pi_{k}(\pos))\cdot\mlp^{L}\left(
    \dir_k, \geometryfeature{k}, \imagefeature{k}(\pi_{k}(\pos))
    \right),
\end{equation}
Note that, to ensure view-consistent light transport estimations, we leverage the multiview-aware feature $\geometryfeature{k}$ in this per-view light transport prediction. 

We repeat this operation to acquire $\ltmat_{1}(\pos), \ltmat_{2}(\pos), ..., \ltmat_{N}(\pos)$. These light-transport functions encode the information at the same 3D point, but are observed from different angles. 
We then calculate final target light-transport $\ltmat(\pos)$ as a linear combination of $\ltmat_{1}(\pos), \ltmat_{2}(\pos), ..., \ltmat_{N}(\pos)$.
We use another MLP $\mlp^{B}$ to predict the blending weights 
\begin{equation}
    \blendingweight{k} = \mlp^{B}\left(
    \dir_k, \dir_t, \geometryfeature{k}
    \right),
\end{equation}
and linearly combine the light-transports of the source views to get the final light-transport
\begin{equation}
    \ltmat(\pos) = \sum_k\blendingweight{k} \odot \ltmat_k(\pos).
\end{equation}

\subsection{Efficient Volume Rendering}\label{sec:render}
For a given target environment map $\envmapmat'$, we follow Equ.~\ref{equ:lt_calc} to compute the radiance $\Mat{L}_r(\pos)$.
The pixel value $\image_t$ of the novel viewpoint can be computed using the volumetric rendering equation from the original NeRF formulation~\cite{mildenhall2020nerf}. Suppose the point at depth $u$ is $\pos(u) = \pos_c + u \cdot \dir_c$, where $\pos_c$ is the camera location and $\dir_c$ is the ray direction, we have:
\begin{equation}\label{equ:vol_render}
    \image_t = \int_{u_{n}}^{u_{f}} \exp\left(-\int_0^u\sigma(\pos(v))\de v\right)\sigma(\pos(u)) \radiance_r(\pos(u)) \de u.
\end{equation}
To be more specific, we integrate along the light ray with $u_{n}, u_{f}$ as the near and far bounds, respectively. We additionally predict the depth and the alpha channel, which are then used for supervision (Sec.~\ref{sec:detail}):
\begin{equation}
\begin{split}
    \Mat{D}_t =& \int_{u_{n}}^{u_{f}} \exp\left(-\int_0^u\sigma(\pos(v))\de v\right)\sigma(\pos(u)) u \cdot \de u,\\
    \Mat{A}_t =& \int_{u_{n}}^{u_{f}} \exp\left(-\int_0^u\sigma(\pos(v))\de v\right)\sigma(\pos(u))  \de u.
\end{split}
\end{equation}

The original NeRF paper computes the radiance at each sample point.
However, most of the sample points are 0 when the scene is spatially sparse. This creates a lot of redundancy when rendering a new image.
In our setup, since portraits are usually sparse, we exploit the concept of \textit{visual hull} to prune out the queries that are unnecessary.
From the input portraits, we utilize a portrait matting algorithm~\cite{ke2020green} to extract their mask $\mask_1, \mask_2, ..., \mask_N$.
Then for each 3D point $\pos$ viewed by the target camera, we project its position onto all the masks, and query the MLPs if all the projections onto $\mask_k(\pi_{k}(\pos))$ are nonzero.
Otherwise, we directly set the light transport $\ltmat(\pos)$ and the density $\density(\pos)$ to be 0.
This is essentially restricting the network to learn within the visual hull defined by the silhouette of the portraits.
By doing this, we achieve faster convergence during training and more efficient rendering during inference.

\subsection{Domain Adaptation}\label{sec:aug}
We train our network on synthetically rendered human face data (see Sec.~\ref{sec:data}). 
This data does not model the distribution of the real human faces very well. 
Thus, naively training our network using the rendered faces results in poor generalizability on real portraits.
It is optimal to include real multi-view portraits in our training data.
However, there is no publicly available large-scale human face dataset and it is also challenging to capture real portraits under different viewpoints and lighting without a light stage.

To this end, we propose a novel domain adaptation module that effectively enhances the generalizability of our network by augmenting the CNN feature extractor with a large number of real portrait images in the CelebAMask-HQ dataset~\cite{CelebAMask-HQ}. 
These real images do not have groundtruth labels of their lighting and camera parameters; however we show that they can be effectively used to regularize our feature extractor to adapt to the distribution of real portrait images. 
In particular, for each training iteration, we additionally feed in a real portrait into the feature extractor and get an image feature.
Rather than using the image feature to predict the light transport and perform volumetric rendering, we append 2 more convolution layers to the U-Net to directly recreate the original image, enhancing the expressiveness of the feature extractor.
With this joint training strategy, we can equip the feature extractor with stronger capabilities to reason about real portrait images and avoid overfitting to the biased distribution of the synthetic portrait dataset.

\subsection{Implementation Detail}\label{sec:detail}
In general, there are two kinds of relighting tasks we need to solve: changing to a completely novel environment light, or rotating the original light.
In order to solve both tasks in NeLF, we implement two modes during training: novel light mode and self rotation mode.
In the novel light mode, we provide a new lighting as the target light to the MLP to compute the radiance.
In the self rotation mode, we predict the lighting of the input portrait from the CNN, rotate the predicted light for a certain angle, and use that as the target light. We allocate 70\% of the training for novel light, and the rest for self rotation.

Our network is supervised with multiple losses.
We implement a rendering loss $\mathcal{L}_c$, a depth loss $\mathcal{L}_d$, an alpha mask loss $\mathcal{L}_a$, a lighting loss $\mathcal{L}_t$ and an image consistency loss $\mathcal{L}_p$.
To begin with, the rendering loss $\mathcal{L}_c$ is simply an $\mathcal{L}_1$ loss on the predicted RGB $\image_{t}$ such that it is as close as possible to the ground truth value $\image_{gt}$
\begin{equation}\label{equ:rendering_loss}
    \mathcal{L}_c = ||\image_{t} - \image_{gt}||_1.
\end{equation}
We additionally supervise the predicted depth and the alpha channel using the ground truth values. This is to make sure the predicted shape is meaningful and obeys multi-view constraints. These losses are formulated as:
\begin{equation}
    \mathcal{L}_a = ||\Mat{A}_t-\Mat{A}_{gt}||_1,
    \mathcal{L}_d = \frac{1}{d}||\Mat{D}_t-\Mat{D}_{gt}||_1, 
\end{equation}
where $d = 200 mm$ is the average size of a human head.
For lighting, we use log-$\mathcal{L}_1$ loss on the predicted environment map $L$ and the ground truth environment map $\envmapmat_{gt}$:
\begin{equation}
    \mathcal{L}_t = ||\log(1+\envmapmat) - \log(1+\envmapmat_{gt})||_1
\end{equation}
As mentioned in Sec.~\ref{sec:aug}, we train our network jointly with a self consistency constraint on the image encoder to ensure that the CNN can learn meaningful representations of real human faces.
For each iteration, we randomly pick a in-the-wild portrait $\image_c$ from the CelebAMask-HQ dataset. We feed the image only into the feature extractor to get a reconstruction of the same image $\hat{\image}_c$. We supervise the reconstruction using an image consistency loss $\mathcal{L}_p$:
\begin{equation}
    \mathcal{L}_p = ||\hat{\image}_c-\image_c||_1.
\end{equation}
Notice that we only apply the image consistency loss on the random in-the-wild portrait $\image_c$. We don't enforce the image encoder to reproduce the calibrated multi-view images. 
Together, the final loss $\mathcal{L}_{total}$ is a sum of all above-mentioned losses:
\begin{equation}
    \mathcal{L}_{total} = \mathcal{L}_c + \mathcal{L}_d + \mathcal{L}_a + \mathcal{L}_t + \mathcal{L}_p.
\end{equation}

The detailed network structures of our CNN and MLP are shown in Fig.~\ref{fig:network}. We implement our method in PyTorch~\cite{NEURIPS2019_9015}. We use Adam~\cite{kingma2014adam} as our optimizer and the learning rates are set to $10^{-4}$ for the MLPs and $2\times10^{-4}$ for the image encoder.
Our network is trained on 4 NVIDIA 2080Ti GPUs for 300k steps, which takes around 36 hours.

    \section{Results}

\begin{figure*}
    \centering
    \begin{tabular}{c@{}c@{}c@{}c@{}c}
    Inputs & Groundtruth & Ours & SIPR+IBRNet & IBRNet+SIPR \\
    \raisebox{-0.5\height}{\includegraphics[width=0.32\linewidth]{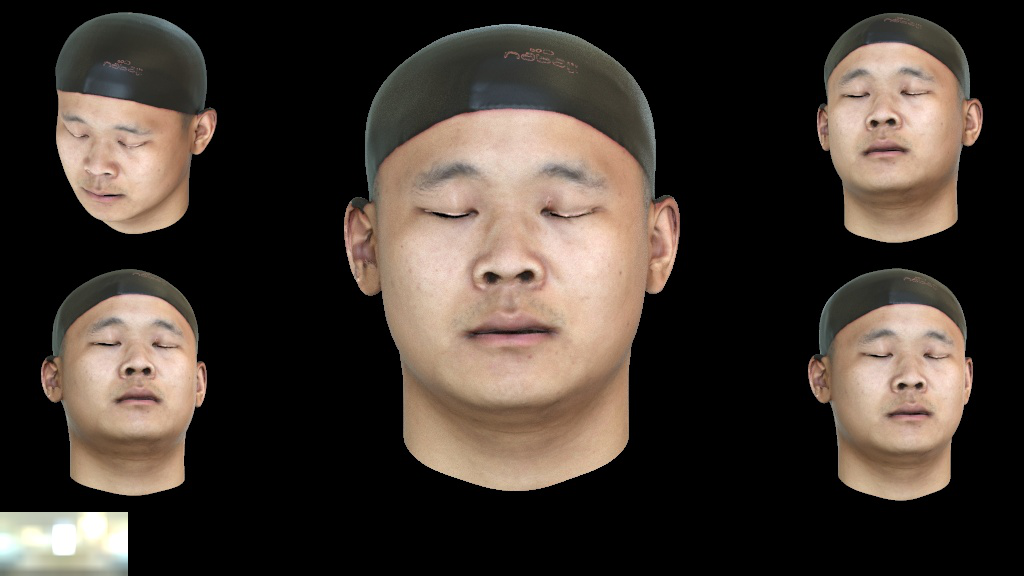}} &
    \raisebox{-0.5\height}{\includegraphics[width=0.16\linewidth]{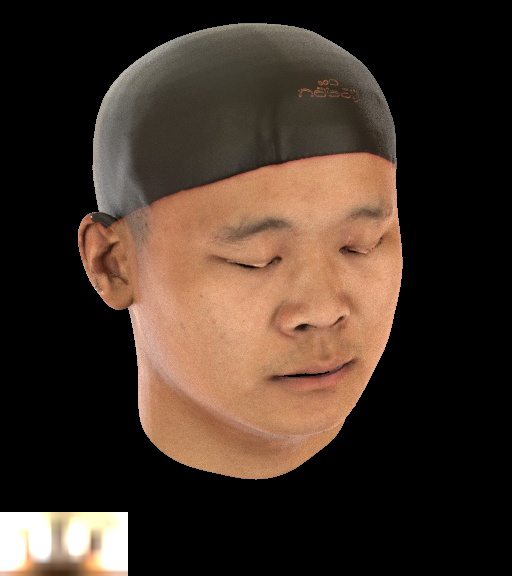}} & 
    \raisebox{-0.5\height}{\includegraphics[width=0.16\linewidth]{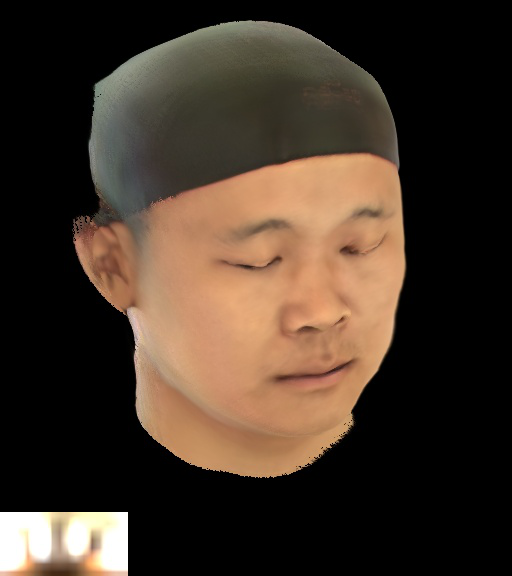}} & 
    \raisebox{-0.5\height}{\includegraphics[width=0.16\linewidth]{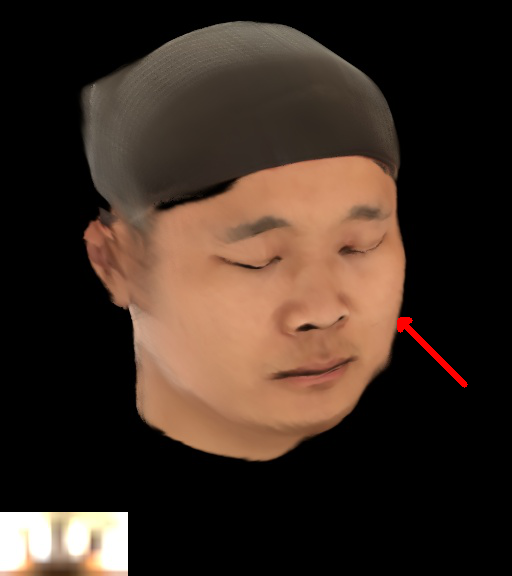}} &
    \raisebox{-0.5\height}{\includegraphics[width=0.16\linewidth]{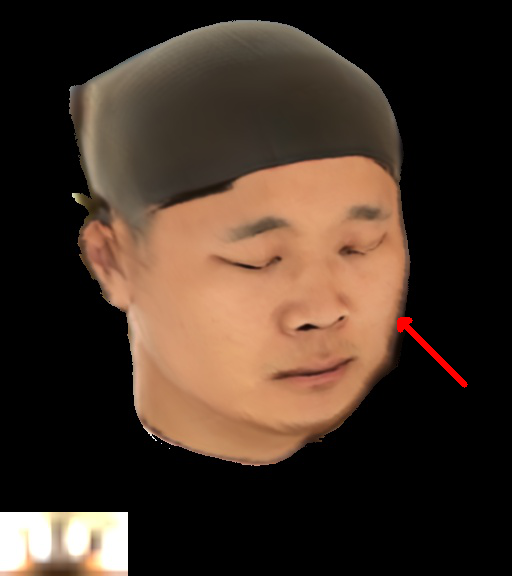}}  \\
    \raisebox{-0.5\height}{\includegraphics[width=0.32\linewidth]{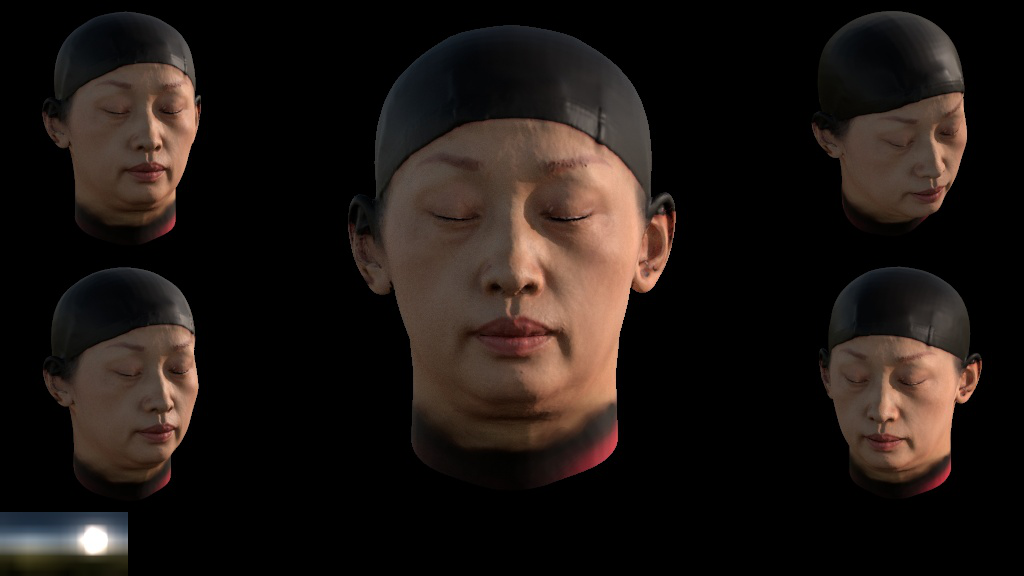}} &
    \raisebox{-0.5\height}{\includegraphics[width=0.16\linewidth]{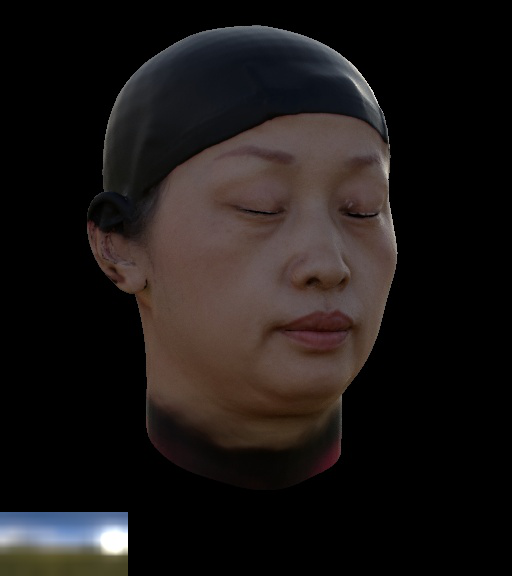}} & 
    \raisebox{-0.5\height}{\includegraphics[width=0.16\linewidth]{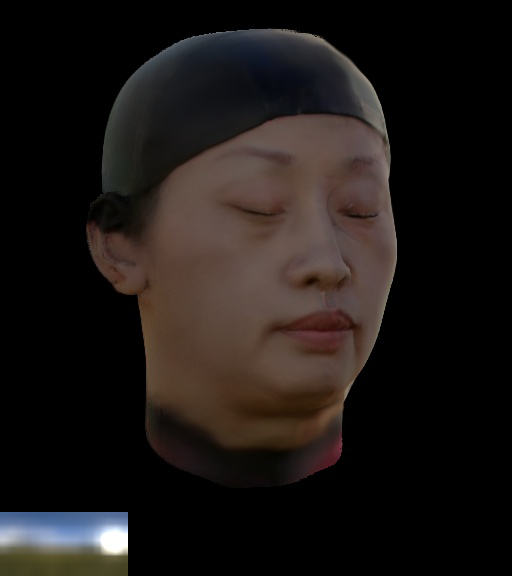}} & 
    \raisebox{-0.5\height}{\includegraphics[width=0.16\linewidth]{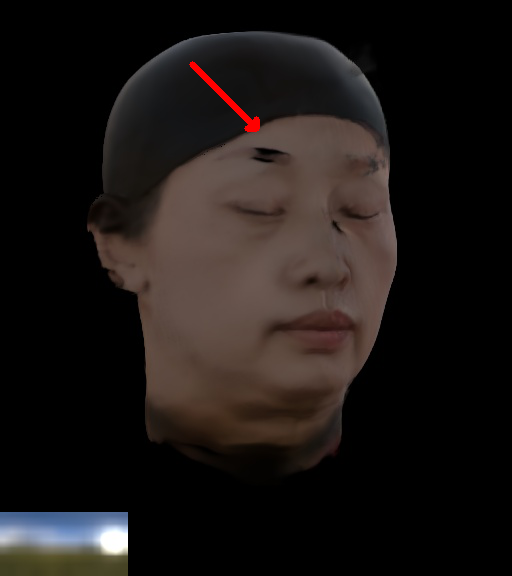}} &
    \raisebox{-0.5\height}{\includegraphics[width=0.16\linewidth]{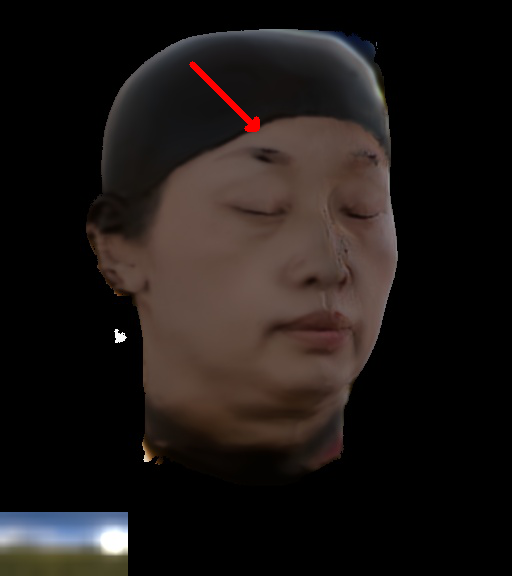}}  \\
    \end{tabular}
    \caption{We compare our results of simultaneous view synthesis and relighting with possible baseline methods on unseen samples from our synthetic evaluation dataset. Baseline methods fail to provide accurate geometry and contain random hole artifacts. For instance, SIPR + IBRNet shows some hole artifacts near the nose area and IBRNet + SIPR shows blurry results around the cheek in the top right image. Please refer to the supplementary video for more results.}
    \label{fig:comp}
\end{figure*}

\subsection{Data}\label{sec:data}

In order to train our novel representation, we choose the FaceScape dataset \cite{yang2020facescape} as it provides a good collection of realistic human head models with high-resolution texture and displacement maps.
Due to privacy issues, some of the models have blurred textures around the eyes.
Thus, we select 360 unblurred models from the whole dataset and use Blender \cite{blender2020} with the Cycles renderer to generate the training and testing data images in 512$\times$512 resolution.
To be more specific, we generate data triplets that contain (a) 5 source views with the first one being the frontal view, (b) a novel view with the same lighting but randomly rotated, and (c) a novel view with lighting randomly selected from a pool of environment maps.
(a) is used as input to our network, (b) is the self rotation mode supervision, and (c) is the novel light mode supervision.
Both (b) and (c) also provide novel view supervision.
The novel views are uniformly sampled within 30$^{\circ}$ in azimuthal and 30$^{\circ}$ in elevational angle from the frontal face view.
The distance of the cameras is chosen randomly from 100cm to 200cm to simulate real life captures.
We also adjust the field-of-view accordingly so that the face would take up most space.
This training triplet design allows for learning of both geometric and lighting information.

\begin{figure*}
    \centering
    \begin{tabular}{c@{}c@{}c@{}c}
    Inputs & Ours & \shortstack{Ours w/\\ direct light transport} & \shortstack{Ours w/o \\domain adaptation} \\
    \raisebox{-0.5\height}{\includegraphics[width=0.38\linewidth]{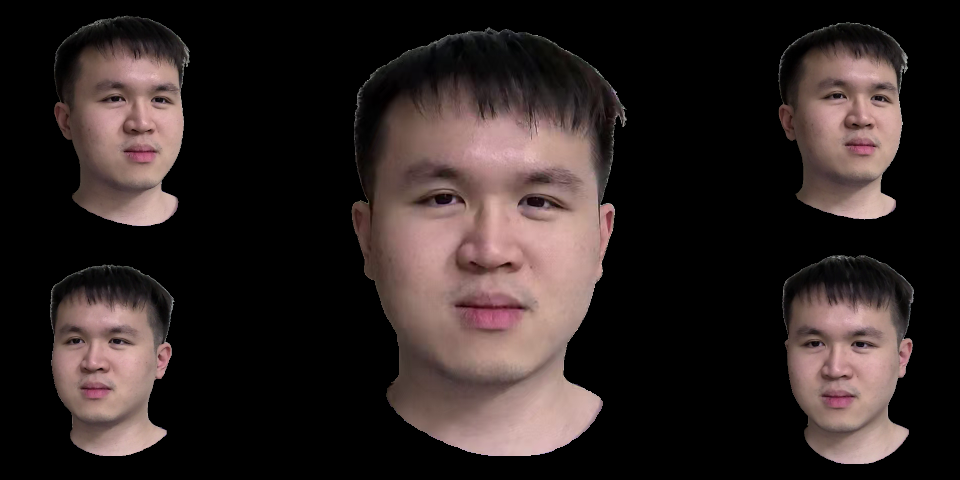}} &
    \raisebox{-0.5\height}{\includegraphics[width=0.19\linewidth]{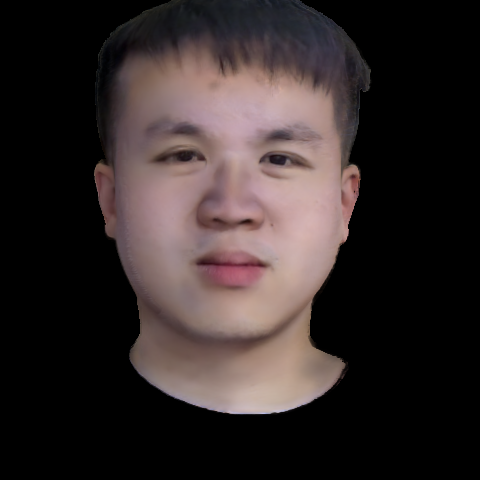}} & 
    \raisebox{-0.5\height}{\includegraphics[width=0.19\linewidth]{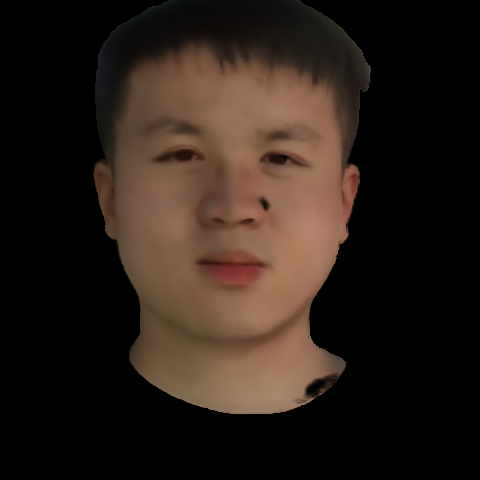}} & 
    \raisebox{-0.5\height}{\includegraphics[width=0.19\linewidth]{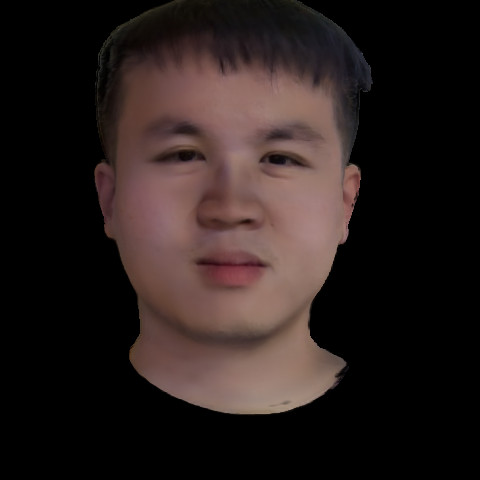}}  \\
    \end{tabular}
    \caption{We compare our method with two possible ablations on a real captured image. We show that by modulating the light transport prediction and applying our domain adaption method, our method can generalize well to real portraits with minimum color shift or artifacts. Please refer to the supplementary video for more results.}
    \label{fig:ablation}
\end{figure*}

\begin{table}[t]
    \centering
    \caption{Quantitative comparison on synthetic evaluation dataset. Our method performs better in both PSNR and SSIM over baseline methods, providing better visual quality. For comparison against direct prediction of light transport, while the visual artifacts are disturbing in Fig. 4, they reflect only a small part of the image, and therefore do not have as much of an impact on the PSNR and SSIM scores. Please refer to supplementary video for better comparisons.}
    \begin{tabular}{|l|cc|}
    \hline
    Method & PSNR & SSIM \\
    \hline
    SIPR + IBRNet & 33.55 & 0.8826 \\
    IBRNet + SIPR & 33.18 & 0.8611 \\
    \hline
    Ours w/o encoder augmentation & 33.54 & 0.8916\\
    Ours w/ direct light transport & \textbf{33.70} & \textbf{0.8928}\\
    \hline
    Ours & \textbf{33.61} & \textbf{0.8922}\\
    \hline
    \end{tabular}
    \label{tab:comp}
\end{table}

\subsection{Comparison with previous methods}\label{sec:comp}
To the authors' best knowledge, our algorithm is the first to achieve simultaneous view synthesis and relighting on unseen subjects from a sparse set of multi-view images under natural illumination. SIPR~\cite{sun2019single} can perform relighting on unseen portraits given a single image, while IBRNet~\cite{wang2021ibrnet} achieves view synthesis on novel scenes. 
We combine these previous two state-of-the-art algorithms and use the combinations as the baseline of our algorithm. There are two possible ways to combine: first do the relighting on the input views, and then synthesize the new view (SIPR+IBRNet); or first do view synthesis and then change the lighting of the synthesized image (IBRNet+SIPR). Notice that we have to correctly rotate the target light to each view in order to align the target light with the camera.

We retrain the network of IBRNet and SIPR on our synthetic dataset for fair comparisons. We choose 4 subjects out of the total 360 identities to serve as our evaluation dataset. Table~\ref{tab:comp} shows the quantitative comparison of these two baselines, as well as our method. Our method outperforms both of the baselines with both higher PSNR and SSIM. In addition, we observe that the baseline methods often lead to obvious visual artifacts in their renderings as shown in Fig.~\ref{fig:comp} (please see our supplementary video for more examples). Our approach instead can achieve much higher visual quality, consistently producing realistic renderings across different lighting and viewpoints.

These two baselines fail for different reasons. If relighting the input images first, the relighting algorithm might perform slightly differently on each input view, which will break the multi-view consistency for view synthesis. On the other hand, doing the relighting after the view synthesis also does not work well, since the relighting algorithm now has no access to the multi-view information. 
In contrast, our approach learns to effectively aggregate multi-view appearance features to predict the light transport of each shading point, leading to accurate relighting effects that are consistent across multi-view viewpoints.
Our approach can simultaneously achieve high-quality relighting and view synthesis.

\begin{table}[t]
    \centering
    \caption{Quantitative comparison on the task of relighting and view-synthesis individually on synthetic dataset. Our method performs comparably, but slightly worse than the state-of-the-art on each task.}
    \begin{tabular}{|l|ccc|}
    \hline
     & Ours & SIPR & IBRNet \\
    \hline
    Relighting     & 0.9053  & \textbf{0.9279} & ---- \\
    View-synthesis & 0.8683  &  ---- & \textbf{0.8949}\\
    \hline
    \end{tabular}
    \label{tab:comp_individual}
\end{table}

\changed{We have also evaluated our method on relighting and view-synthesis respectively using our synthetic dataset. We perform relighting by rendering the view at the frontal camera under new lighting, and do view synthesis by first estimating the original light, and then relighting under the predicted lighting from new views. As shown by the SSIM values in Tab.~\ref{tab:comp_individual}, our method performs comparably, but slightly worse than the baselines on each task. This is because our main goal is not to improve individual performance, but to enable the combination of these two tasks. Thus, our view-synthesis performance also includes the error from lighting estimation and relighting. Even though SIPR and IBRNet perform slightly better in their individual tasks, their combination does not work as well as ours as shown in Fig.~\ref{fig:comp}. Our method aggregates multi-view features to predict the light-transport on each shading point, leading to multi-view consistent relighting effects.}

\begin{figure*}
    \centering
    \begin{tabular}{c@{}c@{}c@{}c@{}c@{}c}
    Inputs & Groundtruth & 5 Views & 4 Views & 3 Views & 2 Views \\
    \raisebox{-0.5\height}{\includegraphics[width=0.28\linewidth]{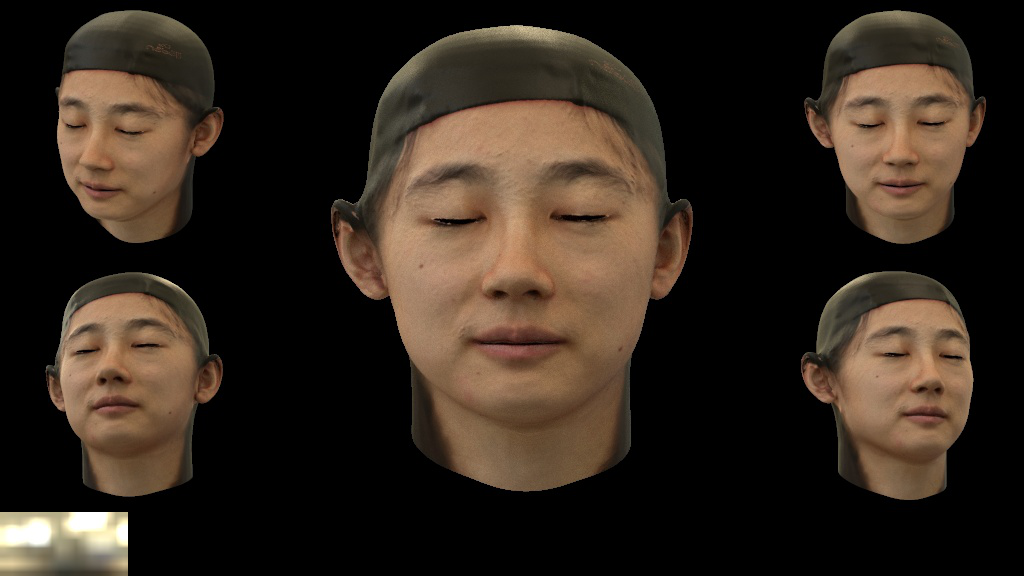}} &
    \raisebox{-0.5\height}{\includegraphics[width=0.14\linewidth]{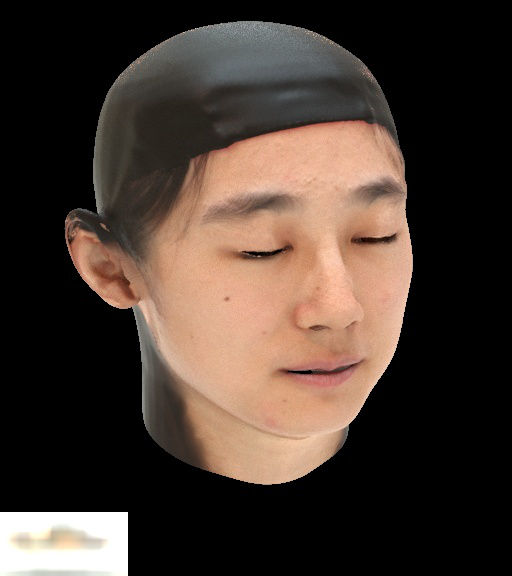}} & 
    \raisebox{-0.5\height}{\includegraphics[width=0.14\linewidth]{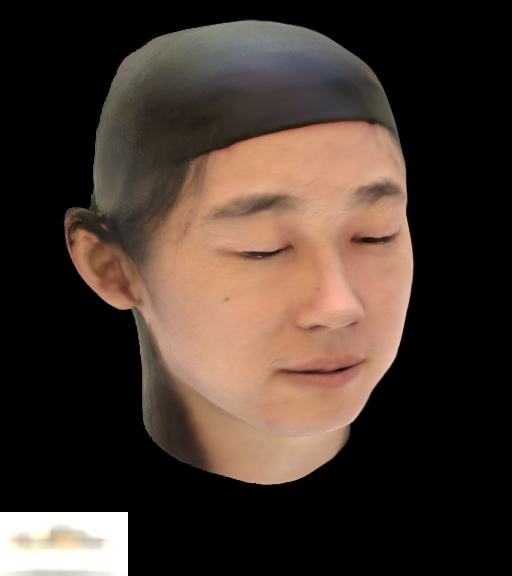}} & 
    \raisebox{-0.5\height}{\includegraphics[width=0.14\linewidth]{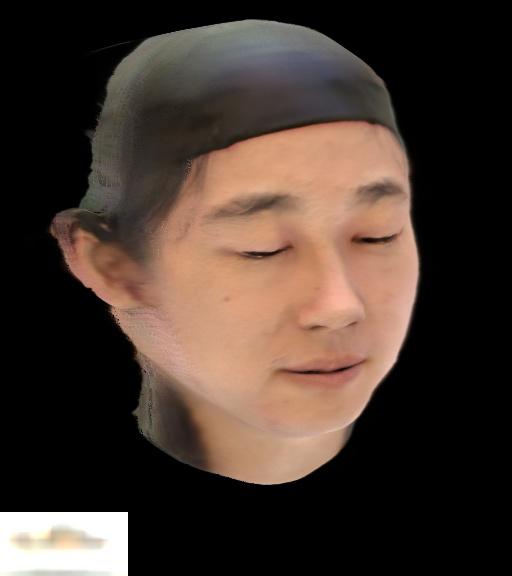}} &
    \raisebox{-0.5\height}{\includegraphics[width=0.14\linewidth]{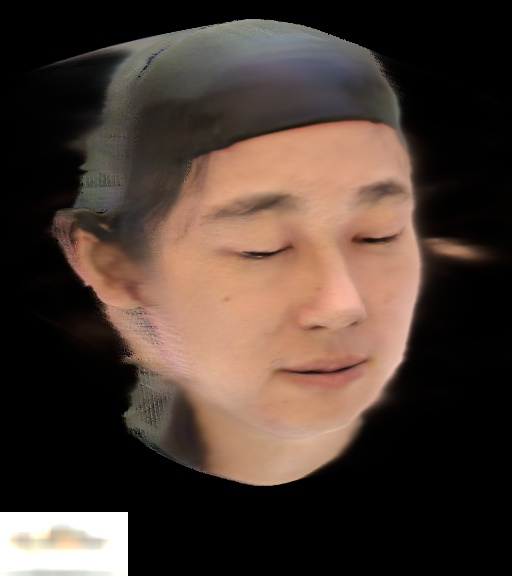}} &
    \raisebox{-0.5\height}{\includegraphics[width=0.14\linewidth]{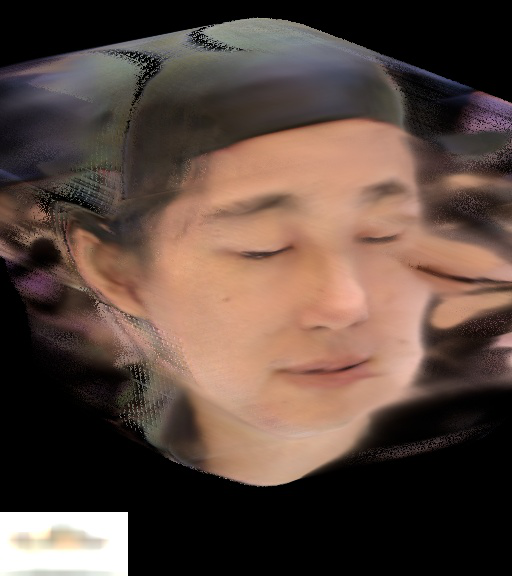}}  \\
    \end{tabular}
    \caption{Qualitative evaluation on the effects of different input view counts. We show that there is a small difference between 4 and 5 views, while 2 and 3 views result in more artifacts due to occlusions and the lack of stereo coverage. }
    \label{fig:view}
\end{figure*}

\subsection{Ablations}\label{sec:abla}
We ablate our algorithm by directly predicting the light transport rather than modulating the prediction using the pixel value in Eqn.~\ref{equ:lt_pred}.
The quantitative results of synthetic validation set and qualitative results of real portraits are shown in Tab.~\ref{tab:comp} and Fig.~\ref{fig:ablation} respectively.
Note that, although the network with direct light transport achieves slightly higher performance on the validation dataset, our full model performs much better on the real portraits as shown in Fig.~\ref{fig:ablation}.
In particular, directly predicting the light-transport can lead to artifacts, for instance the black hole around the nose area in the portrait. In addition, the method introduces more color shift compared to our full algorithm.
Our full model learns to predicts scales of the original pixel colors; this retains the original color signals and can generalize better to unseen real portraits. 

In addition, we show the effect of our domain adaptation module in Fig.~\ref{fig:ablation}.
We show that with the additional adaptation module, our network is able to generalize to unseen color distributions much better than without this module.
As can be seen from the figure, without the adaptation, the network fails to recreate the vibrancy of the input images.
Moreover, the left part of the portrait exhibits more orange color than our proposed method.
Our domain adaptation module essentially regularizes the network to be able to better reproduce the appearance in the original input images. This is not only helpful when testing on real portrait images, but can also improve our performance on the synthetic validation set as shown in Tab.~\ref{tab:comp}.

\begin{table}[t]
    \centering
    \caption{Quantitative comparison on different input view counts. We show that 5 views provide the best possible results in both PSNR and SSIM metrics. Visual quality starts to degrade as view count reduces.}
    \begin{tabular}{|l|cc|}
    \hline
    View Num & PSNR & SSIM \\
    \hline
    2 Views &  30.91 & 0.6388 \\
    3 Views &  32.77 & 0.8177 \\
    4 Views &  33.40 & 0.8731 \\
    \hline
    5 Views & \textbf{33.61} & \textbf{0.8922}\\
    \hline
    \end{tabular}
    \label{tab:view}
\end{table}

We also study the effects of input view numbers on the rendering quality. Our network is originally trained with 5 images as the input views. We test our network on unseen subjects from the validation dataset by feeding 2, 3, and 4 views into the network for view synthesis and relighting. Table~\ref{tab:view} shows the quantitative evaluation and Figure~\ref{fig:view} shows the results. Our network fails to render meaningful content when only 2 images are given. However, as we have 3 or more images as the input views, our network can render new views under new lighting with high quality. 

\begin{figure*}
    \centering
    \begin{tabular}{c@{}c@{}c@{}c@{}c@{}c}
    Inputs & \shortstack{Ours\\View Synthesis} & \shortstack{Ours\\Rotate predict light} & \shortstack{Ours\\Relighting} & SIPR+IBRNet & IBRNet+SIPR \\
    \raisebox{-0.5\height}{\includegraphics[width=0.28\linewidth]{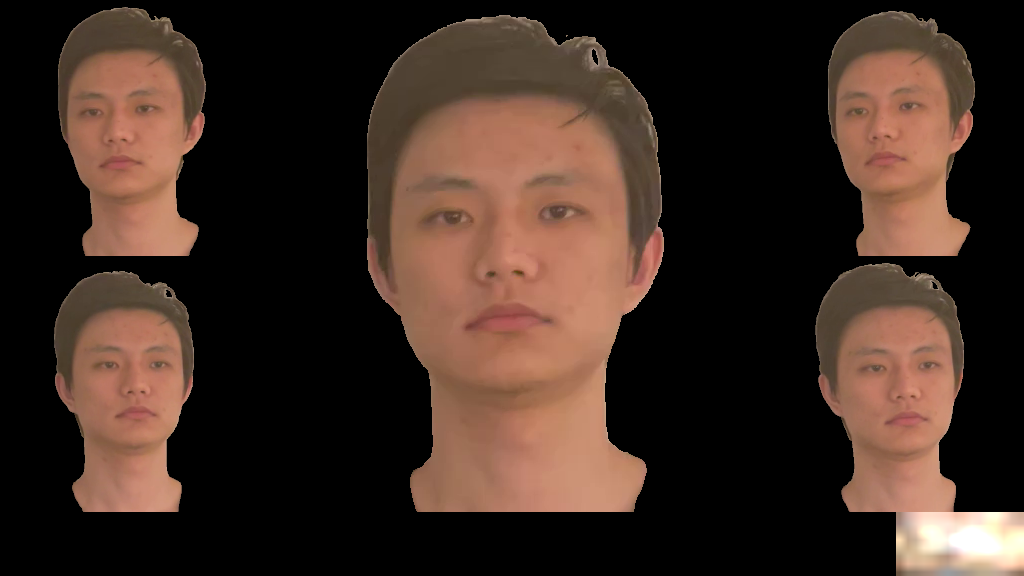}} &
    \raisebox{-0.5\height}{\includegraphics[width=0.14\linewidth]{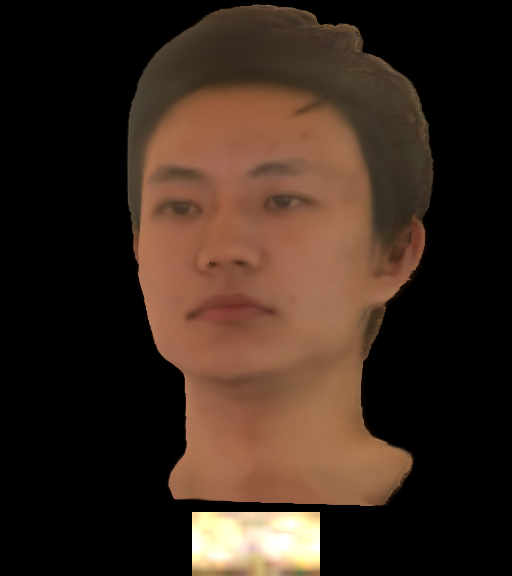}} & 
    \raisebox{-0.5\height}{\includegraphics[width=0.14\linewidth]{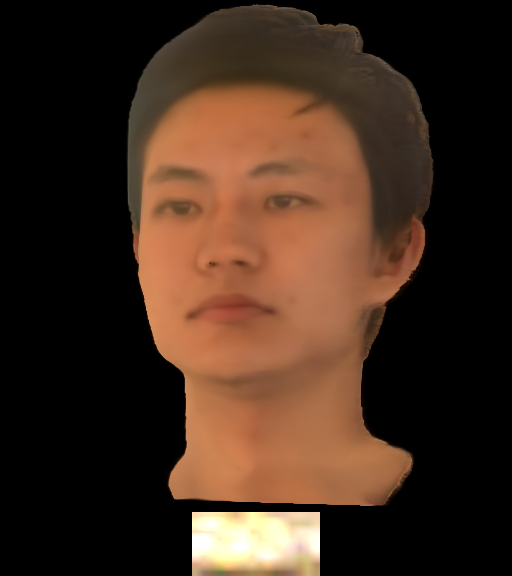}} & 
    \raisebox{-0.5\height}{\includegraphics[width=0.14\linewidth]{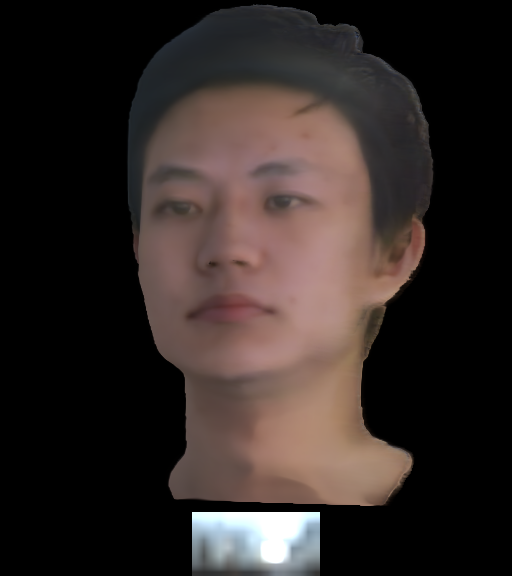}} &
    \raisebox{-0.5\height}{\includegraphics[width=0.14\linewidth]{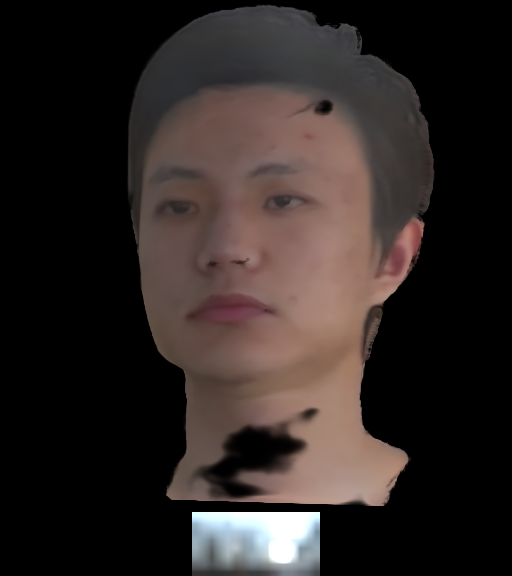}} &
    \raisebox{-0.5\height}{\includegraphics[width=0.14\linewidth]{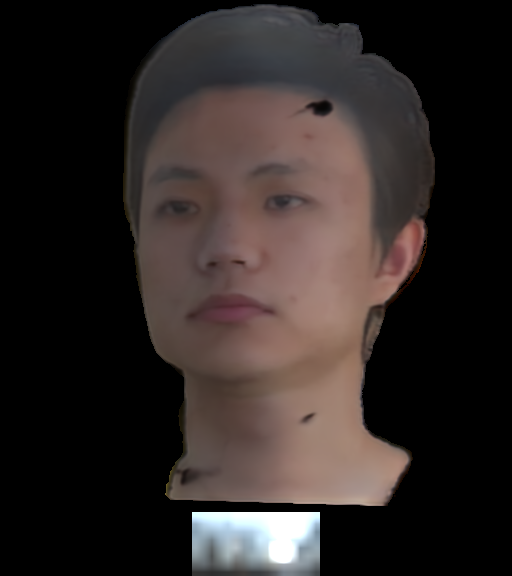}}  \\
    \raisebox{-0.5\height}{\includegraphics[width=0.28\linewidth]{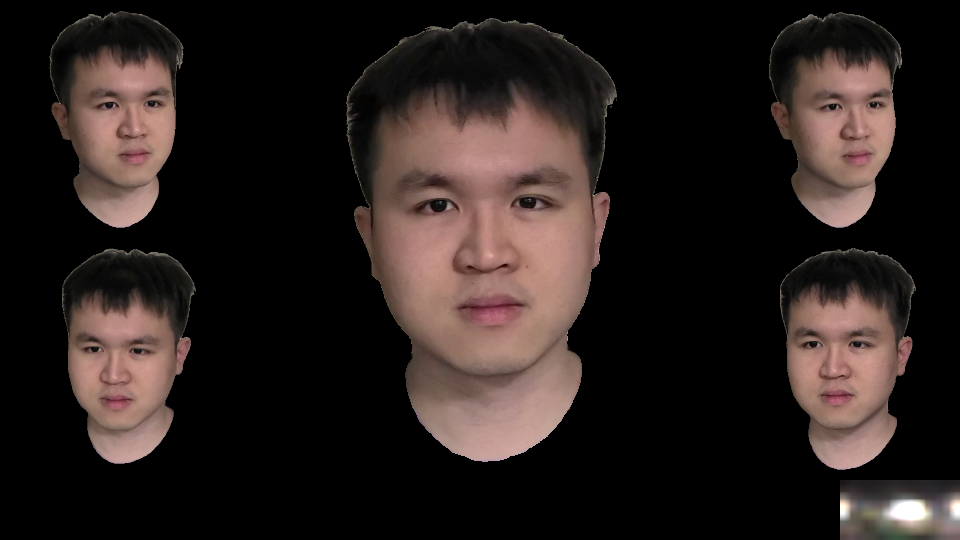}} &
    \raisebox{-0.5\height}{\includegraphics[width=0.14\linewidth]{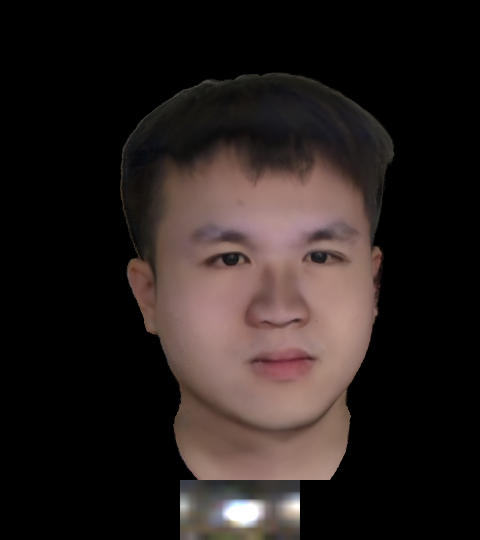}} & 
    \raisebox{-0.5\height}{\includegraphics[width=0.14\linewidth]{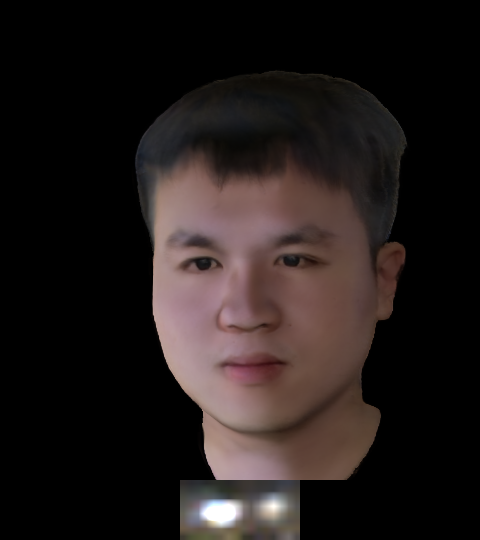}} & 
    \raisebox{-0.5\height}{\includegraphics[width=0.14\linewidth]{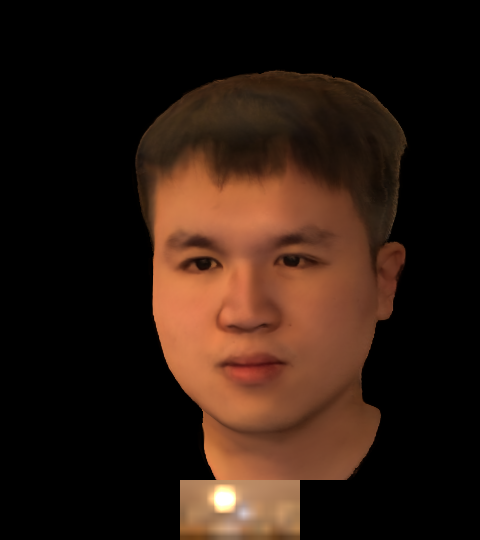}} &
    \raisebox{-0.5\height}{\includegraphics[width=0.14\linewidth]{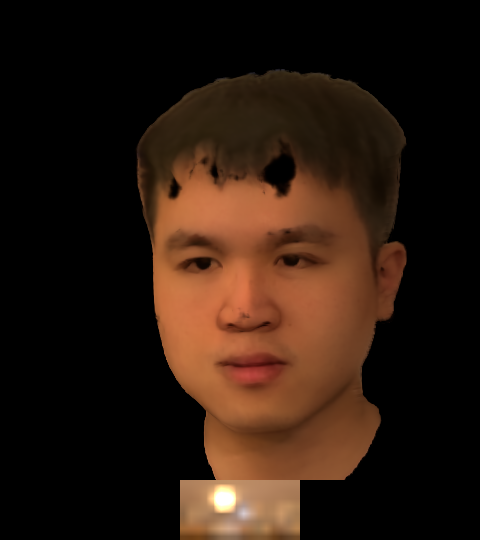}} &
    \raisebox{-0.5\height}{\includegraphics[width=0.14\linewidth]{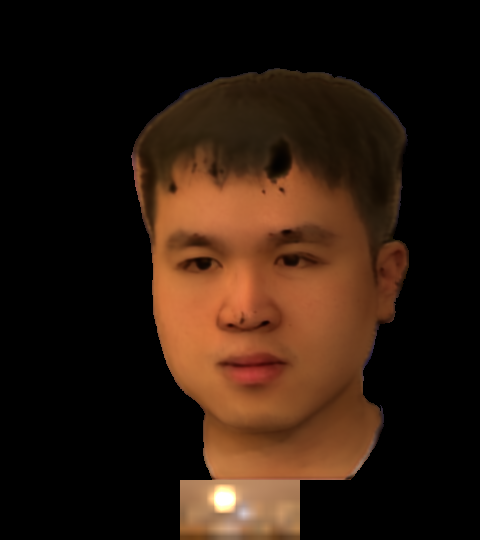}}  \\
    \raisebox{-0.5\height}{\includegraphics[width=0.28\linewidth]{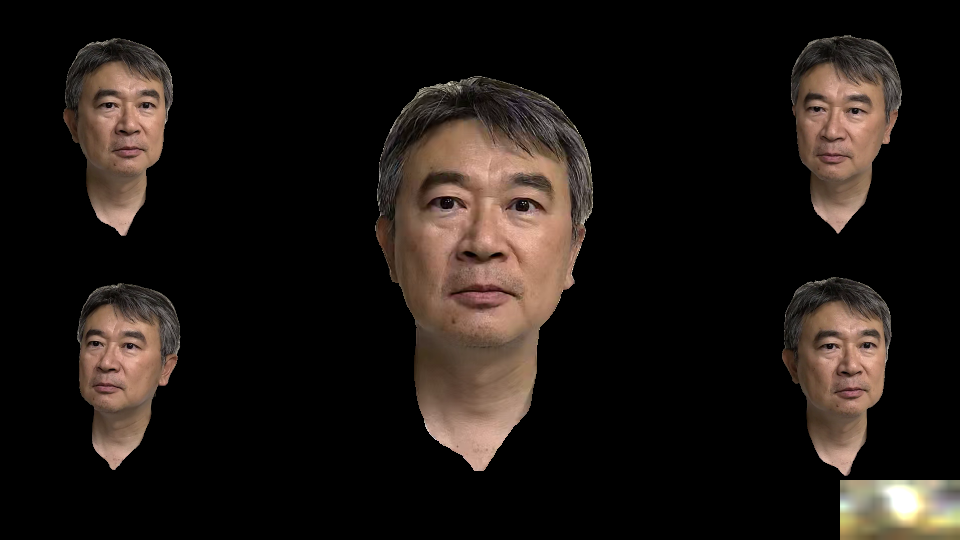}} &
    \raisebox{-0.5\height}{\includegraphics[width=0.14\linewidth]{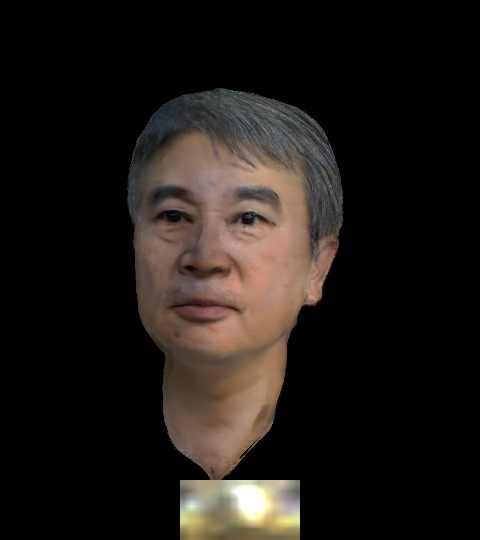}} & 
    \raisebox{-0.5\height}{\includegraphics[width=0.14\linewidth]{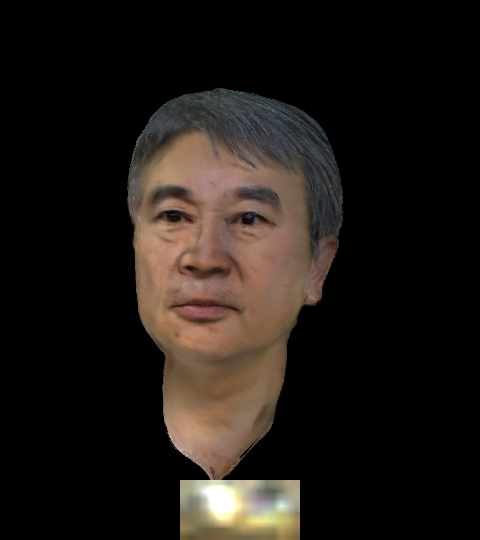}} & 
    \raisebox{-0.5\height}{\includegraphics[width=0.14\linewidth]{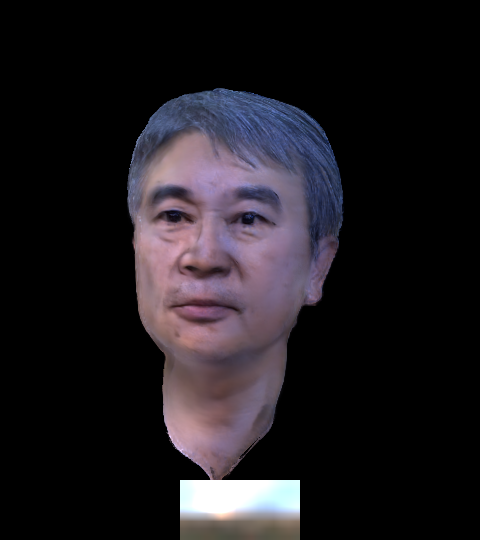}} &
    \raisebox{-0.5\height}{\includegraphics[width=0.14\linewidth]{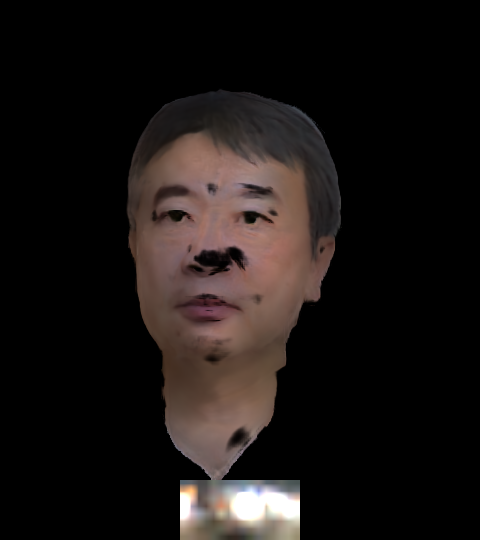}} &
    \raisebox{-0.5\height}{\includegraphics[width=0.14\linewidth]{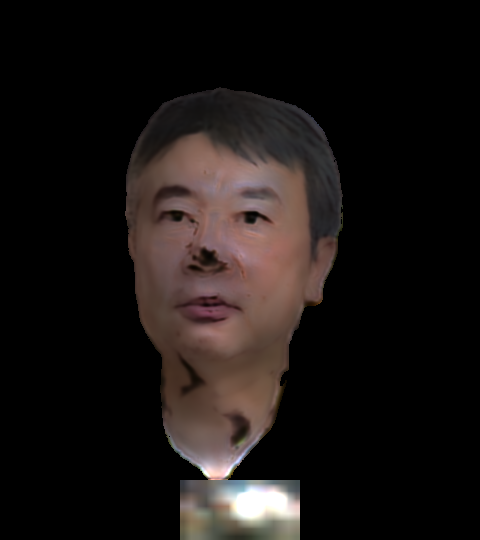}}  \\
    \end{tabular}
    \caption{Qualitative results on real portraits. Our method produces better visual quality than the baseline methods of SIPR and IBRNet. For example, in the forehead and neck regions, baseline methods produce hole artifacts, whereas our proposed method is able to synthesize realistic novel images under new lighting conditions. Please refer to the supplementary video for more results.}
    \label{fig:real}
\end{figure*}

\subsection{Real Portraits}\label{sec:real_portrait}

In order to show the performance of our method on real data, we capture real portraits with smartphones by shooting a video clip of the subject.
We employ COLMAP~\cite{schoenberger2016mvs,schoenberger2016sfm} to recover camera poses and rough depth statistics to determine near and far planes.
We also normalize the camera poses such that the origin is approximately the head center.
This is done by first extracting the center of facial landmarks with Bulat et al.~\cite{bulat2017far}, and estimating the face direction to calculate the head center.
We apply MODNet~\cite{MODNet} to clean up the background and use BiSeNet~\cite{yu2018bisenet} to further remove garments and keep the face portion.
As our training data does not contain any clothes, this can ensure that the rendered results are not perturbed by unrelated information.

The results are shown in Fig.~\ref{fig:teaser}, Fig.~\ref{fig:real} and the supplementary video.
We can observe that our method achieves state-of-the-art quality on the joint task of view synthesis and relighting.
Our proposed method is able to infer source environment map and render novel view portraits with the rotated source light.
Additionally, we can input a target environment map to control the new lighting condition.
For the task of rendering real portraits, the comparison baselines (both SIPR+IBRNet and IBRNet+SIPR) lead to obvious artifacts with many visible holes, more obvious than their artifacts on the synthetic validation set; this is because the baseline methods' networks overfit to the training set's data distribution and the same issue of inconsistency between separate relighting and view synthesis modules (as discussed in Sec.~\ref{sec:comp}) becomes more significant on the real data.
Thanks to our effective modeling of the light transport in the 3D space and our domain adaptation module, our model can synthesize realistic relighting and view synthesis results on the real portrait images, leading to significantly better results than the baseline methods.

\subsection{Limitations}\label{sec:limits}
Although our proposed method generates photo-realistic rendering results, it still possesses some limitations.
For example, since we train on a synthetic dataset, our model might exhibit some color shifts in certain cases when tested on real portraits.
This can be ameliorated by training on a real multi-view portrait dataset.
Another issue is the slight blurriness in our rendered results.
This is possibly caused by the limited network capacity, which can be further increased to allow extraction of image features with higher resolution.
In addition, we downsample the feature map to allow for faster training and inference.
With more computational resources, it is possible to use full-scale image features.
Last, because our training dataset have a global specularity coefficient instead of dedicated specularity maps, some complex view-dependent effects such as glints are not well 
reconstructed in the training images.
As a result, our method fails in some cases where high-frequency specular highlights are presented.

    \section{Conclusions and Future Work}
In this paper, we tackle the joint problem of portrait view synthesis and relighting, which prior works fails to handle.
We solve this problem by introducing the \textit{neural light-transport field}, which encodes the volume density and light transport vectors of each 3D point in the scene, enabling relighting with a target environment map.
We demonstrate that with only 5 input views, our method is able to generalize across unseen portraits and produce better portrait renderings than previous approaches built specifically for either view synthesis or relighting.

In future work, we would like to explore the possibility of even fewer input images. We show that our method degrades for a smaller number 
of views.
This is caused by the depth ambiguities and unseen areas, which requires prior knowledge to hallucinate reasonable renderings.
Another possible direction is animated relightable avatars as our proposed method does not handle a talking head explicitly.
All in all, we believe that the joint problem of view synthesis and relighting is crucial in immersive applications like mixed reality, and our work can advance the field in that direction.
    
    \section*{Acknowledgement}
    This work was supported by ONR grant N000142012529 and N000141912293, NSF grant 1730158, a Google Ph.D. Fellowship, a Qualcomm FMA Fellowship, an Amazon research award and gifts from Adobe.
    Thanks to the reviewers for the valuable feedback, and to the anonymous volunteers for being captured.

    \printbibliography

\end{document}